\documentclass{article}

 \usepackage[nonatbib,final]{nips_2016}
\usepackage{graphicx}
\usepackage{color}

\usepackage[utf8]{inputenc} 
\usepackage[T1]{fontenc}    
\usepackage{url}            
\usepackage{booktabs}       
\usepackage{amsfonts}       
\usepackage{nicefrac}       
\usepackage{microtype}      
\usepackage{amsmath,bm}
\newcommand{\bx}{\boldsymbol{x}}

\newcommand{\bw}{\boldsymbol{w}}
\newcommand{\ba}{\boldsymbol{a}}
\newcommand{\bs}{\boldsymbol{s}}

\title{Investigating the influence of noise and distractors on the interpretation of neural networks }

\author{
  Pieter-Jan Kindermans\\
 Machine Learning Group\\
  TU Berlin\\
  \texttt{p.kindermans@tu-berlin.de} \\
  \And
    Kristof T. Schütt\\
 Machine Learning Group\\
  TU Berlin\\
  \texttt{kristof.schuett@tu-berlin.de} \\
  \And
     Klaus-Robert Müller\\
 TU-Berlin\\
  Korea University\\
  \texttt{klaus-robert.mueller@tu-berlin.de} \\
  \And
    Sven Dähne\\
 Machine Learning Group\\
  TU Berlin\\
  \texttt{sven.daehne@tu-berlin.de} \\
}

\begin{document}

\maketitle

\begin{abstract}
Understanding neural networks is becoming increasingly important. Over the last few years different types of visualisation and explanation methods have been proposed. However, none of them explicitly considered the behaviour in the presence of noise and distracting elements. In this work, we will show how noise and distracting dimensions can influence the result of an explanation model. This gives a new theoretical insights to aid selection of the most appropriate explanation model within the deep-Taylor decomposition framework.
\end{abstract}
\section{Introduction}

In recent years, deep learning had a huge impact on the machine learning community beating the state of the art across a wide range of applications (e.g. ImageNet classification~\cite{krizhevsky2012imagenet}).
Beyond applying deep learning to novel tasks and improving performance, considerable effort has been put into making machine learning models more interpretable.
One of the first works on this topic, Zeiler et al.~\cite{Zeiler2014} propose the Deconvnet as a means to visualize the features learned by the neural network.
Simonyan et al.~\cite{Simonyan2013} propose class-specific saliency maps corresponding to the gradient of an output neutron with respect to the input.
Both approaches are related and differ mainly in the positioning of the ReLu units~\cite{Simonyan2013}.
Bach et al.~\cite{Bach2015} define relevance as the contribution of each input variable to the prediction introducing layer-wise relevance propagation (LRP). 
Montavon et al. \cite{Montavon2015} provide a theoretical basis for LRP and generalize the method to the deep Taylor decomposition.
It is striking that, while neural networks explanation methods are mainly based on the gradient,
such a procedure is considered unsuitable to obtain an interpretation in Neuroimaging due to noise and distractor signals in the data~\cite{Haufe2014}.
This motivated the analysis of explanation methods below.

\section{Linear projections in the presence of noise and distractors}

We start with the following assumptions based on~\cite{Haufe2014}:
The observed data $\bx$ is generated by a model $\bx=\ba_{t}s_{t}+A_{n}\bs_{n}^{T}+\mathbf{\epsilon}$, being composed of a task-related component $\ba_{t}s_{t}$ and a noise component $A_{n}\bs_{n}^{T}+\mathbf{\epsilon}$.
Here, $s_{t}$ is the hidden signal to be recovered, i.e., the class label, regression output or hidden neuron activation prior to the non-linearity.
The vector $\bs_{n}$ comprises uncorrelated latent signals with $E \left[\bs_{n}\right]=\mathbf{0}$ and $Cov[s_t,\bs_{n}]=\mathbf{0}$, $\mathbf{\epsilon}$ contains zero mean Gaussian noise. 
While $s_{t},\mathbf{s_{n}}$ and $\mathbf{\epsilon}$ are example-specific, the patterns $\ba_{t},A_{n}$ are shared across the data set.
The vector $\ba_{t}$ describes how changes to the task-related signal result in changes to our observed data $\bx$.
The matrix $A_{t}$ describes in which directions the data can change without changing the desired output. Note that there is no requirement that $\ba_{t}$ and the components in $A_{n}$ are orthogonal to each other. 
A linear projection $\bw$, trained to detect a specific feature or class membership indicated by $s_{t}$, has the following expected output:
\[
E\left[\bw^{T}\bx\right]=E\left[\bw^{T}\left(\ba_{t}s_{t}+A_{n}\bs_{n}^{T}+\mathbf{\epsilon}\right)\right]=s_{t}.
\]
Hence, it is required that $\bw^{T}\ba_{t}=1$ such that the desired target signal $s_{t}$ can be recovered. 
Given $\ba_{t}$, this can be fulfilled by infinitely many $\bw$. 
Additionally, $\bw$ has to be orthogonal to task-unrelated variations: $\bw^{T}A_{n}=\mathbf{0}$.
This is critical to the interpretation of $\bw$.
The weight vector of a linear projection yields the direction of steepest ascent concerning the output of the classifier.
It is important to realize that this is not necessarily the task-related direction of variation because of the orthogonality requirement between $\bw$ and $A_n$.
For this reason, gradient-based approaches, like deconvnet and saliency maps, are known to be uninformative for linear classifiers in noisy environments \cite{Haufe2014} (e.g. EEG analysis).
On top of that, the reasoning above makes clear that we implicitly introduce additional assumptions about our data and networks when using different explanation rules or visualisations.

\section{What assumptions do explanation rules make implicitly?}

{\bf The deep taylor decomposition} \cite{Montavon2015} performs Taylor expansions in a layer-wise fashion distributing the relevance to the inputs of the neural network\footnote{For simplicity we will assume that biases are introduced by a constant input neuron with activation $1$. It is easy to show that in this setting, for ReLu non-linearities and max-pooling, the saliency map multiplied elementwise with the input corresponds to the explanation obtained by the $z$-rule. This proof is included in the appendix}.

The goal of the deep taylor decomposition is to decompose the output of a neural network into contributions that can be assigned to the different input variables (i.e. pixels when considering images). 
Formally this is described as follows. Let $R^\text{out}_j=f_i(\bx)$ be the $j$th output of the neural network for $D$-dimensionl input $\bx=[x_1,\ldots,x_D]^T$. We want to have a decomposition of $R^\text{out}_j=\sum_{i=1}^D R^{in}_i$ where we sum over the $D$ inputs of the neural network.   The key idea being that inputs pixels that contribute more to the final output of the network are more important than pixels that contribute less. 

The goal of decomposing the output into input contributions leads to the requirement as an additional constraint, the total relevance in each layer stays constant. 
The relevance of an output neuron $j$ is considered to be $R_{j}^{\text{out}}=x_{j}$.
The upper-layer relevance is redistributed employing a first degree Taylor decomposition around a root point $\tilde{\bx}^{j}$ which is chosen to fulfill $\bw_{j}^{T}\tilde{\bx}^{j}=0$.
The relevance of layer $l$ is redistributed to the neurons of the previous layer following 
$
\sum_{j} R_{j}^l=\sum_{i} \sum_{j}\left.\frac{\partial R_{j}^l}{\partial x_{i}^{l-1}}\right|_{\tilde{x}_{i}^{j}}.\left(x_{i}^{l-1}-\tilde{x}_{i}^{j}\right)+\eta = \sum_{i} R_{i}^{l-1} + \eta,
$
where $R_{i}^l$ and $x_{i}$ denote the relevance and activation of neuron $i$ in layer $l$, respectively.
The residual $\eta$ vanishes for ReLu non-linearities.
This redistribution is applied recursively to obtain the relevance of each input feature.


The selection of a root point can be done based on different search
directions. 
In the original paper, a search direction is chosen based
on constraints of the domain of the previous layer, e.g., if the input neuron is limited in range or constrained to be positive as it is the output of a ReLu layer. 
In contrast, we argue to think about the search direction of the root point with respect to the generative model introduced above instead of the input domain.
Ideally, we find the direction in which the data shows task-related variations.
In the remainder of this section, we will take a look at how various propagation rules for deep Taylor decomposition correspond to different generative models defined by $\ba_{t},A_{n}$ and propose two new rules.
Not all known explanation rules are considered due to space constraints.

{\bf The $z$-rule} was originally proposed in \cite{Bach2015} and later integrated in the
deep Taylor framework \cite{Montavon2015}.
It corresponds to a root point which is a
constant $\mathbf{0}$, yielding $\bx=\ba_{t}s_{t}+A_{n}\bs_{n}^{T}$
with $\ba_{t}=\frac{\bx}{\bw^{T}\bx}$ and $A_{n}\bs_{n}=\mathbf{0}$.
Hence, when using the z-rule, we assume that noise is not present in
our data and that all input directions are informative\footnote{Previous work by Montavon
et al. \cite{Montavon2011} has shown that the amount of noise decreases as one goes to
higher layers of the neural network. Consequently, this rule can be
used reliably for decomposing higher level neurons but one should
be careful when applying it to lower layer neurons on noisy data.}. 


{\bf The $w^{2}$-rule and a new $w^{+}$-rule}
The $w^{2}$-rule was proposed for neurons with unbounded domain (i.e., the input layer).
This rule chooses the closest root point lying along the direction of the gradient.
In the generative model this corresponds to $\ba_{t}=\frac{\bw}{\bw^{T}\bw}$ and  $A_{n}\bs_{n}=\bx-\frac{\bw}{\bw^{T}\bw}s_{t}$.
While it was originally intended to be used for the input layer, we argue that the $w^{2}$-rule can be applied across the entire network with a slight modification: the search direction is now chosen such that  relevance can only be propagated to neurons with non-zero activation: $\bw^{+}=\bw\odot\mathbf{1}_{\bx\neq0}$. 
This yields $\ba_{t}=\frac{\mathbf{w^{+}}}{\mathbf{w^{+}}^{T}\mathbf{w^{+}}}$ and assumes that the detected features lie along the gradient with respect to the input in the active input neurons. 
The assumption that the data varies along the direction of the gradient is also used by the saliency map and the deconvnet.

{\bf The $a$-rule and $a^{+}$ rule.}
Finally, we propose a new rule based on what is commonly applied in Neuroimaging~\cite{Haufe2014}. We will learn the vector $\ba_{t}$ by regressing from the output of the layer (before non-linearity) to each input neuron. We make the simplifying assumption that all neurons are independent. Thus, the task-relevant direction is approximated for each neuron as: $\hat{\ba}=(\bw^T XX^T\bw)^{-1} \bw^TX X^T,$
where $X$ contains the inputs to the neuron. The matrix $X=[\bx_1,\ldots,\bx_N]$ contains one data point $\bx_n$ per column. In the $a$-rule, we use this direction directly. As a result, the explanation would not be data-point dependent. Therefore, we make it adaptive by limiting the directions to the active input neurons and rescaling in the $a^+$-rule if it was preceeded by a ReLu activation.

\section{Experiments and Discussion}
\begin{figure}[t]
\centering
\includegraphics[width=\textwidth]{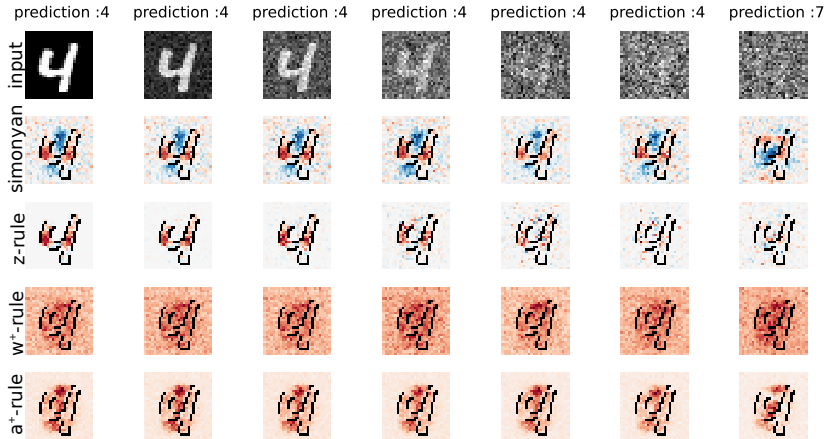}
\caption{Comparison of the behaviour of various explanation methods under the influence of noise. See text for details.\label{fig:decoding}}
\end{figure}
We present a preliminary evaluation of the introduced rules based on the classification of digits on the MNIST dataset with an MLP.
The images are scaled in the range $[0,1]$ and normal noise has been added with a standard deviation ranging 0.0 to 0.8. 
The MLP possesses a hidden layer with 200 fully connected neurons and ReLu non-linearity and a softmax output layer.
As suggested by Simonyan et al.~\cite{Simonyan2013} and Bach et al.~\cite{Bach2015}, the explanation is started from the output layer before applying the softmax.
We compare the saliency map and the $z$-rule, to the new  $w^+$ and $a^+$ rule we proposed here. 

We look at Figure: ref{fig:decoding} first. This figures highlights the influence of increasing noise levels on the different relevance methods.
The saliency map, being essentially the gradient, is mimicking the structure of the input. 
While there is some considerable background noise in the explanation, it remains quite stable when the noise level is increased. Indicating that the network keeps using the same input pixels. As predicted, the $z$-rule is very crisp when no noise is present. This is partially caused by the inputs being zero since it is equal to the saliency map multiplied with the input. However, this also causes the quality of the heatmap to degrade when the noise level is increased. This is not surprising since the implied, generative model assumes that there is no noise present.

Compared to $z$-rule, the explanation of the $w^+$ and $a^+$ rules are much less inpacted as the noise level increases. The $w^+$ rule stands out by assigning positive relevance to the entire input image. On the other hand, the $a^+$ method focusses more on the center region of the image where there is most variation.
This difference is caused by the gradient, which has non-zero weights over the entire input image.
The learned $\ba$ vectors show only the directions with class-relevant variation in the data.
We argue that for MNIST focussing on the center is reasonable since this is where the input images differ and class information is present.

From a qualitative point, we observe in this setting that both Simonyan's saliency map approach and the $a^+$-rule indicate that the gaps between the two vertical lines at the upper part as well as the bottom of the four are important. This intuitively makes sense since this would change the four into a nine or eight, respectively. 

\begin{figure}
\centering
\includegraphics[width=\textwidth]{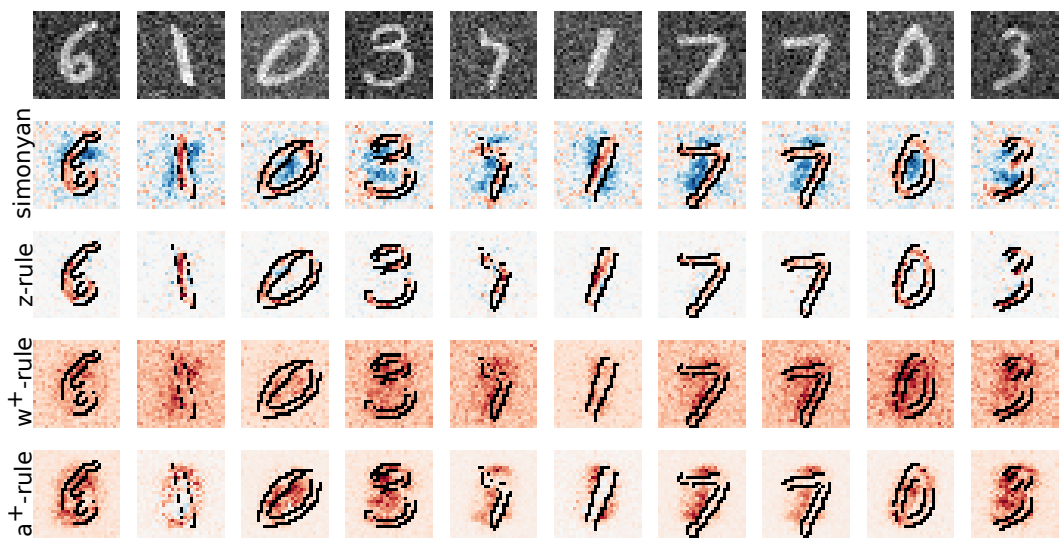}
\caption{Comparison of heatmaps on different MNIST digits with the noise level set to 0.2. See text for details.\label{fig:decoding2}}
\end{figure}

In Figure: \ref{fig:decoding2} we present more images from the MNIST dataset. 
Here we observe a general trend. The $z$-rule focusses on the symbol and assigns almost no relevance to the region outside of the actual number. 
The Simonyan saliency map and the two other decomposition rules assign much relevance to the region outside of the actual number. For the $a^+$-rule this effect is particularly strong and the visualisation indicates that the black regions next to the number itself are more important than the actual number.

\section{Conclusion}
In this abstract, we have taken a different theoretical view on the explanation of neural networks using the deep taylor decomposition.
We have shown how different explanation rules correspond to different generative models of our observed data and how this influences the explanation. 
Based on our observations, we have also introduced a new explanation rule that learns the directions of variation related to the task.
While this abstract has generated new insights into the explanation process, further research on large networks and additional datasets is needed to pick the appropriate model for each task and network architecture.
Furthermore, benchmark taks where the relevant component is known are needed to make the evaluation of relevance visualisations on neural networks more reliable.

\subsubsection*{Acknowledgments}
This project has received funding from the European Union's Horizon 2020 research and innovation programme under the Marie Sklodowska-Curie grant agreement NO 657679.
Additional support was provided by the Federal Ministry of Education and Research (BMBF) for the Berlin Big Data Center BBDC (01IS14013A).

\bibliographystyle{plain}
\bibliography{LRP_NOISE}

\begin{thebibliography}{1}

\bibitem{Bach2015}
Sebastian Bach, Alexander Binder, Gr{\'e}goire Montavon, Frederick Klauschen,
  Klaus-Robert M{\"u}ller, and Wojciech Samek.
\newblock On pixel-wise explanations for non-linear classifier decisions by
  layer-wise relevance propagation.
\newblock {\em PloS one}, 10(7):e0130140, 2015.

\bibitem{Haufe2014}
Stefan Haufe, Frank Meinecke, Kai G{\"o}rgen, Sven D{\"a}hne, John-Dylan
  Haynes, Benjamin Blankertz, and Felix Bie{\ss}mann.
\newblock On the interpretation of weight vectors of linear models in
  multivariate neuroimaging.
\newblock {\em Neuroimage}, 87:96--110, 2014.

\bibitem{krizhevsky2012imagenet}
Alex Krizhevsky, Ilya Sutskever, and Geoffrey~E Hinton.
\newblock Imagenet classification with deep convolutional neural networks.
\newblock In {\em Advances in neural information processing systems}, pages
  1097--1105, 2012.

\bibitem{Montavon2015}
Gr{\'e}goire Montavon, Sebastian Bach, Alexander Binder, Wojciech Samek, and
  Klaus-Robert M{\"u}ller.
\newblock Explaining nonlinear classification decisions with deep taylor
  decomposition.
\newblock {\em arXiv preprint arXiv:1512.02479}, 2015.

\bibitem{Montavon2011}
Gregoire Montavon, Mikio~L Braun, and Klaus-Robert M{\"u}ller.
\newblock Kernel analysis of deep networks.
\newblock {\em Journal of Machine Learning Research}, 12(Sep):2563--2581, 2011.

\bibitem{Simonyan2013}
Karen Simonyan, Andrea Vedaldi, and Andrew Zisserman.
\newblock Deep inside convolutional networks: Visualising image classification
  models and saliency maps.
\newblock {\em arXiv preprint arXiv:1312.6034}, 2013.

\bibitem{Zeiler2014}
Matthew~D Zeiler and Rob Fergus.
\newblock Visualizing and understanding convolutional networks.
\newblock In {\em European Conference on Computer Vision}, pages 818--833.
  Springer, 2014.

\end{thebibliography}
\clearpage

\appendix

\section{The connection between the $z$-rule and the Simonyan saliancy map.}
In this section, we will show that the $z$-rule with biases included as input neurons is equivalent to the gradient (Simonyan's saliency map) multiplied elementwise with the input. We are assuming that there is only a ReLu activation or max-pooling. The effect of non-active ReLu neurons and paths blocked by max-pooling is included taking the sum only over active neurons.

\subsection{Proof}
By definition, the relevance for a neuron in the top layer $t$ is defined as follows:
\[
R_{i}^{t}=x_{i}^{t}.
\]
The relevance of the layer below is: 
\begin{eqnarray*}
R_{j}^{t-1} & = & \sum_{i}\underbrace{R_{i}^{t}}_{x_{i}^{t}}\frac{w_{j}^{t-1,i}x_{j}{}^{t-1}}{\underbrace{\sum_{j}w_{j}^{t-1,i}x_{j}^{t-1}}_{x_{i}^{t}}}\\
 & = & \sum_{i}x_{i}^{t}\frac{w_{j}^{t-1,i}x_{j}{}^{t-1}}{x_{i}^{t}}\\
 & = & \sum_{i}w_{j}^{t-1,i}x_{j}{}^{t-1}.
\end{eqnarray*}
This can be expressed as the gradient multiplied with the neuron activation of the previous layer:
\[
R_{j}^{t-1}=\frac{\partial x_{i}^{t}}{\partial x_{j}^{t-1}}x_{j}^{t-1}.
\]
Given that for layer $t-n$ the $z$-rule explanation is
\[
R_{j}^{t-n}=\frac{\partial x_{i}^{t}}{\partial x_{j}^{t-n}}x_{j}^{t-n},
\]
then for layer $t-n-1$ holds:
\begin{eqnarray*}
R_{k}^{t-n-1} & = & \sum_{j}R_{j}^{t-n}\frac{w_{k}^{t-n-1,j}x_{k}^{t-n-1}}{\underbrace{\sum_{k}w_{k}^{t-n-1,j}x_{k}^{t-n-1}}_{x_{j}^{t-n}}}\\
 & = & \sum_{j}\frac{\partial x_{i}^{t}}{\partial x_{j}^{t-n}}x_{j}^{t-n}\frac{w_{k}^{t-n,j}x_{k}^{t-n-1}}{\underbrace{\sum_{k}w_{k}^{t-n-1,j}x_{k}^{t-n-1}}_{x_{j}^{t-n}}}\\
 & = & \sum_{j}\frac{\partial x_{i}^{t}}{\partial x_{j}^{t-n}}w_{k}^{t-n,j}x_{k}^{t-n-1}\\
 & = & \sum_{j}\frac{\partial x_{i}^{t}}{\partial x_{j}^{t-n}}\frac{\partial x_{j}^{t-n}}{\partial x_{k}^{t-n-1}}x_{k}^{t-n-1}\\
 & = & \frac{\partial x_{i}^{t}}{\partial x_{k}^{t-n-1}}x_{k}^{t-n-1}.
\end{eqnarray*}
This shows that the $z$-rule is equivalent to the saliency map multiplied with the input.
\subsection{Implications}
Since the gradient of the weight vector for a network with only ReLu and max-pooling units yields a linear projection that performs the identical operation for this specific datapoint as the neural network itself.  Hence, the fact that the $z$-rule can be decomposed into the gradient multiplied with the input proves that for these networks the $z$-rule results in a valid decomposition of the actual function the neural network computes. However, this decomposition, as discussed before, assumes that there is no (structured) noise or distracting pattern present in the data. This approach sees the neural network as a data-point specific function generator.
\end{document}